\documentstyle[aaai]{article}
\title{Implementing Integrity Constraints\\ 
in an Existing Belief Revision System\thanks{This work was supported in part by the US Army
Communications and
Electronics Command (CECOM), Ft. Monmouth, NJ through a contract with CACI
Technologies.}
\\
CSE Technical Report 2000-03
}

\author{ Frances L. Johnson \and  Stuart C. Shapiro \\ Department
of Computer Science and Engineering \\ and Center for Multisource Information
Fusion and Center for Cognitive Science \\ State University of New York at
Buffalo\\ 226 Bell Hall,  Buffalo, NY  14260-2000, USA \\  
\textit {\{flj $|$ shapiro\}@cse.buffalo.edu} 
}
\date{March 7, 2000}

\begin{document}
\maketitle

\begin{abstract}
\noindent 
SNePS is a mature knowledge representation, reasoning, and acting
system that has long contained a belief revision subsystem, called
SNeBR.  SNeBR is triggered when an explicit contradiction is
introduced into the SNePS belief space, either because of a user's new 
assertion, or because of a user's query.  SNeBR then makes the user
decide what belief to remove from the belief space in order to restore
consistency, although it provides information to help the user in
making that decision.  We have recently added automatic belief
revision to SNeBR, by which, under certain circumstances, SNeBR
decides by itself which belief to remove, and then informs the user of 
the decision and its consequences.  We have used the well-known belief 
revision integrity constraints as a guide in designing automatic
belief revision, taking into account, however, that SNePS's belief
space is not deductively closed, and that it would be infeasible to
form the deductive closure in order to decide what belief to remove.
This paper briefly describes SNeBR both before and after this
revision, discusses how we adapted the integrity constraints for this
purpose, and gives an example of the new SNeBR in action.
\end{abstract}

\section{Introduction}
Belief revision, or belief change, is the term used to describe any change in a
knowledge base.  The form of belief revision discussed in this paper is removal
of propositions from a knowledge base that is known to be inconsistent in order
to restore consistency.  This is especially important to information fusion,
where information is combined from multiple sources which might contradict each
other.  This paper describes some belief revision theories and the
considerations that arose when they were implemented and added to the
existing belief revision subsystem of a mature 
 knowledge representation and reasoning system,
SNePS\cite{ShaSIG99a,ShaRap91a}.

These considerations center around the impossibility of implementing deductive
closure and the proper weighting of belief revision guidelines. 
We address the need to formalize theories that take into account the fact that
deductive closure cannot be guaranteed in a real-world, need-based, implemented
system. We also explore one technique for following the belief revision
guideline of minimizing damage to the belief space while retaining the most
credible beliefs.

The next section provides the background necessary to understand the alterations we
made to SNePS.  Included are brief descriptions of the following: (a) the
different groups doing belief revision research, (b) the integrity constraints
we intend to implement, (c) SNePS and its belief revision sub-system, (d) the
status of constraint adherence before system alterations, and (e) previous
research to improve adherence. 

The following two sections discuss the current changes made to improve adherence and
implement 
automatic belief revision (autoBR).  The latter gives the user a sense of
non-monotonicity, because it is possible to add a belief to the belief space and,
consequently, lose a previously held belief. However the underlying
relevance-style, paraconsistent logic remains monotonic.

The final section contains conclusions and plans for future work.

\section{Background}

\subsection{Theory vs. Implementation}
Belief revision research can be divided into two groups:
Theoretical vs. Implementations.  These two groups differ in the amount of
information  assumed to be in a knowledge base. The theoretical
researchers develop postulates about how a knowledge base should react during
revision based upon a number of guidelines.  One of these is the assumption of
deductive closure, which means that everything derivable is contained in the
knowledge base---a deductively closed belief space (DCBS), which is infinite in
size and impossible to implement, although it can be simulated with
limitations.

Implemented belief spaces must be finite and of reasonable size, and the
reasoning operations performed on them must be done in a finite and reasonable
time.  Beliefs must be added gradually over time and not all derivable beliefs
can be guaranteed to be present in the knowledge base. We refer to this kind of
belief space as
a deductively open belief space, or DOBS\cite{JohSha00a}.  In a DOBS, it is
possible for a
proposition to be derivable from the existing belief space without it being
present
in that belief space.  
If that proposition is the negation of another proposition derivable from the
belief space, the belief space would be inconsistent,
 but not known to
be
inconsistent---i.e. the system would be unaware of the inconsistency.  The normal
meaning of the word \emph{consistent} cannot, therefore, be applied to a DOBS.
For
the purpose of this paper, however, the term consistent will be used to
describe a belief space or belief set that has no known contradictions (unless
otherwise noted).

Even research on finite belief bases still refers to deductive
closure
in its determination of consistency or when considering belief revision
postulates \cite{Han93a,Han93b,Neb89,AlcMak85}. Consistency is determined by the presence
of a contradiction in the \emph{implicit} beliefs of a belief base. How can this
reliably be implemented? Even if it can be simulated on a small scale knowledge
 base, the theories developed might be suspect if applied to a very large
system. Nebel voiced these concerns as well \cite{Neb89}.

We address the need to formalize theories that take
into account the fact that deductive closure cannot be guaranteed in a
real-world, need-based, implemented system---even if the system is restricted
by something as simple as the user needing a response within one minute. These
theories need to suggest a belief revision technique that:
\begin{itemize}
\item takes time and complexity limitations into account
\item recognizes that adhering to these limitations might result in revision
choices that are poor in hindsight
\item catches and corrects these poor choices as efficiently as possible.
\end{itemize}

\subsection{Integrity Constraints for Belief Revision}
G\"{a}rdenfors and Rott \cite{GarRot95} discuss postulates for belief revision using a
coherence approach. These postulates, first presented in \cite{AlcGarMak85}, 
are based on four integrity constraints (paraphrased below):
\begin{enumerate}
\item a knowledge base should be kept consistent whenever possible;
\item if a proposition can be derived from the beliefs in the knowledge base, then it should be included in that knowledge base;
\item there should be a minimal loss of information during belief revision;
\item if some beliefs are considered more important or entrenched than others, then belief revision should retract the least important ones.
\end{enumerate}

These constraints come from the Theorist group, so strict adherence to
constraints 1 and 2 in an implemented system is impossible.    The proper
weighting and combining of constraints 3 and 4 remains an open question in
belief revision research and a challenge for both the Theorists and the
Implementers.

\subsection{SNePS and SNeBR Before Alteration}
\subsubsection{SNePS}
SNePS is a logic- and network-based knowledge representation, reasoning, and
acting system designed to constitute the mind of a natural language competent
cognitive agent. The
underlying logic of SNePS is a monotonic, relevance-style, paraconsistent
logic \cite{MarSha88}.  One way that users can interact with SNePS is through the SNePSLOG
interface---an interface which allows the user to input propositions in a style
that uses ``predicate calculus augmented with SNePS logical connectives''
\cite{MckMar81,ShaSIG99a}---where
propositions are expressed as well-formed formulas, or wffs.  

Propositions added to the knowledge base by the user are called
hypotheses. Propositions derived from those existing in the belief space are
called derived propositions.
The system records a justification for each proposition (whether it is a
hypothesis or derived) by associating it with an \emph{origin set} consisting of the
hypotheses used in its derivation. An origin set for a belief is a set of
hypotheses which is known to \emph{minimally derive} that
belief---i.e. no subset of a belief's
origin set is known to derive that belief. 

This is along the style of an ATMS, short for
``assumption-based'' truth maintenance system ---a term ``introduced by \cite{deK84,deK86}, although
similar ideas had been investigated earlier by \cite{MarSha83}.'' \cite{Mar90}
also mentions that \cite{San67} presented the ``first description of as
ATMS-like system.''

A hypothesis has a singleton
origin set, containing only itself, but might also be derivable from other
hypotheses.  Multiple derivations of a single proposition can result in its
having multiple origin sets.  If the hypotheses in a proposition's origin set
are asserted, or believed, then the proposition is also believed and is part of
the belief space. This situates SNePS firmly on the foundations side of the
coherence/foundations belief revision divide, but a coherence approach can be
simulated by additionally asserting each derived belief as a hypothesis.

In the SNePS terminology, the belief space is the set of believed
propositions---both hypotheses and derived beliefs---which are supported by the
current context.  The current context is, intensionally, a named structure 
that  contains a set of hypotheses. That set is the extensional context. When 
a new hypothesis is added, the intensional context now contains a different
extensional set of hypotheses. When we refer to adding and
removing hypotheses from ``the context'', we are referring to the intensional
context.

Unlike theoretical knowledge bases, implemented ones cannot promise deductive
closure for a knowledge base, because of the space and time limitations of the
real world.  SNePS attempts to derive propositions as they are asked for---either by the user or by the system as it performs backward chaining or forward
inference. This is typical of a DOBS as it is formalized in \cite{JohSha00a}.

\subsubsection{SNeBR}
The SNePS belief revision sub-system, SNeBR \cite{MarSha88}, is
activated when a derived proposition or a hypothesis is added to the belief space,
\emph{and} it explicitly contradicts a pre-existing belief. The following are examples
of explicit contradictions in SNePS:
\begin{itemize}
\item \verb|P|  and \verb|~P|
\item \verb|P|  and \verb|~(P| $\vee$ \verb|Q)|
\end{itemize}
but \emph{not} 
\begin{itemize}
\item \verb|Q| and \verb|Q=>P| and \verb|~P| 
\end{itemize}
because, in this last case, \verb|P| is only an implicit belief and must be derived before that contradiction can be detected.
 
The detection of an explicit contradiction is almost instantaneous, even in a
knowledge base with thousands of nodes, due to the Uniqueness Principle \cite{MaiSha82}, which
states that no two SNePS
terms denote the same entity.  \begin{quote}[Therefore an]
explicit contradiction \ldots  in the belief space, is easily
recognized by the system because \ldots  the data structure representing \texttt{P} is directly
pointed to by the negation operator in the data structure representing
\verb|~P|. \cite{ShaJoh00}\end{quote}
The user has the option of letting the
belief space remain inconsistent or activating a manual version of belief revision
to restore consistency.  The latter is performed by forming a
minimally-inconsistent set of hypotheses, which can be made consistent upon the
removal of any one of its members.  This set is the union of the origin sets
for the contradicting beliefs---multiple origin sets for a belief result in
multiple inconsistent sets to be revised.  For example: If the contradictory
propositions \verb|P| and \verb|~P| had one ($\alpha$) and two ($\beta$
and $\gamma$) origin sets
respectively, then there would be two minimally-inconsistent sets formed for
belief revision: ($\alpha$ $\cup$ $\beta$) and ($\alpha$
 $\cup$ $\gamma$).  

After forming the inconsistent sets, SNeBR prompts the user to remove at least
one proposition from each set to restore consistency.  It is up to the user to
decide which beliefs should be removed (retracted, become unasserted). 
 
\subsection{How SNeBR Adheres to Constraints 1 and 2}
\subsubsection{Adherence to Constraint 1}
Because a SNePS belief space is a DOBS, it can only claim consistency in terms of
not knowing of any contradictions.  Immediately upon discovery of a
contradiction, however, the system activates SNeBR to restore consistency.
Therefore, the user always has the option to maintain consistency ``whenever
possible''. 

\subsubsection{Adherence to Constraint 2}
Although SNePS cannot promise that all derivable beliefs are in the knowledge
base, it does derive a proposition upon query if that proposition is derivable
from the existing knowledge base.  Therefore, SNePS follows an altered version
of constraint 2: If a proposition can be derived from the beliefs in the
knowledge base, then the system will produce it if it is asked for. 

\subsection{Previous Attempts at Ordering Beliefs and AutoBR}
The researchers described in this section chose to order beliefs based on
relative credibility \emph{as determined by the user}---i.e. the user decides which
beliefs (or types of beliefs) are more credible than others.  This not the only way to order beliefs as ``more
important or entrenched,'' but, due to our interest in information fusion, this epistemic entrenchment is the way we, also,
have chosen to order our beliefs.  To this end, any reference in this paper to
orderings of beliefs or sources should be assumed to mean ordering based on
credibility (unless otherwise stated).  
This ordering was then used in the
implementation of automatic belief revision (autoBR) which allowed the systems described to
perform belief revision without user interaction.

Cravo and Martins \cite{CraMar93a} introduced an altered version of SNePS, called SNePSwD
(SNePS with Defaults),  that incorporated default reasoning.  It also offered
automatic belief revision based on ordering beliefs by credibility and
specificity.  The system allowed the user not only to order beliefs but to also
order the orders. Ordering large amounts of information was tedious, however,
and any new additions required updating relevant orderings.

Ehrlich \cite{Ehr95,EhrRap97,RapEhr2000}
altered a version of SNePSwD that had this ordering capability.  She chose to
eliminate the laborious hand ordering by defining a predetermined group of
 ``knowledge categories,''  with a preset ordering.  
The category for a proposition
was added as another argument of the proposition.  She then ordered all her
propositions based on the relative order of their knowledge categories.
Although this saved her the manual ordering, there were several drawbacks:  (a)
she had to predetermine the knowledge categories that would be used, (b) no new
ones could be added, and (c) the ordering hierarchy of the categories was
fixed. 
 
Both Cravo and Martins and Ehrlich developed their automatic belief revision
processes (autoBR) to remove hypotheses based on credibility orderings---adhering to constraint 4.  How to properly combine and weight both constraints
3 and 4 remains an open topic in belief revision.  Our initial attempt to
combine and weight them is detailed later.

\section{Alterations to Aid Adherence to Constraint 3}
To minimize information loss, the total number of beliefs removed from the
system during revision must be considered. Removal of some belief, \verb|P|,
will also remove any propositions that have \verb|P| in all of their currently active
origin sets.  The
revised SNeBR system orders the hypotheses in the inconsistent sets based on
the number of derived propositions that they support.  

It is also possible to
make multiple inconsistent sets consistent with a single retraction in the
case that a
hypothesis common to those sets is the one chosen for removal.  The revised
system creates an ordered list of the hypotheses based on how many of the
minimally-inconsistent sets they are in.  

From these two orderings, two lists
are formed: (a) the hypotheses supporting the fewest number of derived propositions,
and (b) the hypotheses common to the largest number of inconsistent sets. These
two lists are now considered during belief revision in the interest of
adherence to constraint 3. 

Unlike belief spaces created by deductive closure, our DOBS system builds
beliefs as they
are queried about, thus we can consider those beliefs to be of high interest to
the user. This somewhat validates the connection between the cardinality of a
belief set and the information it holds. Other researchers  \cite{SimKuc98} also choose culprits based on set cardinality
combined with credibility issues.

Our current approach to minimizing information loss involves counting
believed propositions without regard to their internal form.  For
example, \verb|A| and \verb|B|$\wedge$\verb|C|$\wedge$\verb|D| are each counted as one proposition, as are
both \verb|Q(a)| and \verb|all(x)(P(x) => Q(x))|, even thought the latter one of
each pair clearly "encodes" more information.  Especially in the
context of a DOBS, it would be important to try to assess the
usefulness of a proposition in terms of the number of other
propositions that might in the future be derived from it.  We leave
this assessment for future work.

\section{Epistemic Entrenchment Additions to SNeBR}
\subsection{Sources and their information}
As mentioned earlier, Ehrlich's knowledge categories have to be determined and
ordered off-line before running the system. This, plus the source information
being included in each proposition as additional argument, forced a static
treatment and implementation. Our revised SNePS system uses meta-propositions
to assign sources, allowing dynamic source addition and ordering.

\subsubsection{Problems with Source Information as an Added Argument}
Representing the source information as an additional argument of the predicates has several problems associated with it.  For example, if ``Fran is smart.'' were represented as \verb|Smart(Fran)|, ``The prof says that Fran is smart.'' could be represented as \verb|Smart(Fran, Prof)|, and the problems are:
\begin{enumerate}
\item The source of \verb|Smart(Fran, Prof)| cannot be removed or changed without also removing or changing the belief that Fran is smart.  Although that might immediately be reintroduced with \verb|Smart(Fran, Nerd)|,  belief revision may have had to be performed in the interim, wasting time and effort.
\item The proposition \verb|~Smart(Fran, Sexist)| might represent either the belief that the sexist is the source of the information that Fran is not smart or the belief that the sexist is not the source of the information that Fran is smart.  No matter which one it does represent, there is no obvious way to represent the other.
\item It is not clear how to ascribe a source to a rule, such as	
\verb|all(x)(Grad(x) => Smart(x))|.
\end{enumerate}

\subsubsection{Benefits of Source Information in a Meta-Proposition}
Representing the source information in a meta-proposition, is to represent it
as a belief about the belief.  For example, ``The prof says that Fran is
smart.'' would be represented as \verb|Source(Prof, Smart(Fran))|.\footnote{This is
syntactically and semantically correct, because propositions like
\verb|Smart(Fran)| are, in SNePS, functional terms denoting propositions \cite{Sha93a}.}  This solves the three problems cited above:
\begin{enumerate}
\item The source of the belief that Fran is smart can be removed or changed, without removing or changing the belief that Fran is smart, by removing \verb|Source(Prof, Smart(Fran))| without removing \verb|Smart(Fran)|, and then, perhaps, introducing a different source, e.g. \verb|Source(Nerd, Smart(Fran))|.
\item The belief that the sexist is the source of the information that Fran is not smart would be represented as \verb|Source(Sexist, ~Smart(Fran))|, whereas the belief that the sexist is not the source of the information that Fran is smart would be represented as \verb|~Source(Sexist, Smart(Fran))|.
\item The belief that the prof is the source of the rule that all grads are smart would be represented by      
\verb|Source(Prof, all(x)(Grad(x) => Smart(x)))|.
\end{enumerate}

As shown, this allows dynamic interaction, where the user can add, remove, and
change source information about a proposition while the system is running and
without affecting or entirely rewriting the actual proposition.  Source
orderings are also stored as propositions, such as  \verb|Greater(Prof,Nerd)|, which
can be interacted with and reasoned about dynamically.  New sources as well as
their credibility orderings can be added at any time to the knowledge base.
Propositions can even have more than one source, although SNeBR assumes a
single source at this time.

\subsection{Ordering Propositions (Epistemic Entrenchment) and Sources}
Our system currently depends on the user to determine the credibility of
sources or beliefs directly. The user inputs information to the system
declaring source credibility orders and/or belief credibility orders. These are
partial orders that are qualitative and transitive. 

The system currently assumes beliefs have, at most, one
source, but future research will explore multiple source situations and their
implementation. We are also assuming at this time that a more credible source
delivers more credible information. E.g. Given the following: 
\begin{itemize}
\item Lisa is more credible than Bart. 
\item Lisa tells us, ``It is snowing.'' 
\item Bart tells us, ``Homer is fat.''
\end{itemize}
we consider ``It is snowing'' more credible than ``Homer is fat.''

Since ordering can never be assumed complete or unchangeable, the system works
with what it has---including propositions whose sources are unknown (assumed at
this time to be more credible than beliefs that have recorded sources).
An
interesting issue to explore in the future would be to have the system
dynamically
establishing and adjusting source credibility information based on 
revision experiences. 

\subsection{Recommendations and Automatic Belief Revision}
The user can set the belief revision mode at any time from the top level of the
SNePSLOG interface.   The two modes are:
\begin{description}
\item[manual] 		offers recommendations, but requires the user to revise (this is the default)
\item[auto]		activates automatic belief revision, autoBR
\end{description}

From the union of all the inconsistent sets underlying the contradiction, SNeBR produces three lists that are used to create a recommended culprit list:
\begin{description}
\item[LB] 	the least believed hypotheses
\item[MC]    	the hypotheses that are the most common  (to the largest number of inconsistent sets)
\item[FS]    	the hypotheses that support the fewest beliefs in the knowledge
base.
\end{description}

The first set provides possible culprits that support constraint 4 (remove the
least important or least credible beliefs).  Both the second and the third sets
will provide possible culprits that support constraint 3 (minimal loss of
information during belief revision).  The culprit list is created by combining
these lists using a method described in the next section.

Once the recommended culprit list is formed, the user is notified of the three
lists as well as the culprit list.  In manual mode, the user must then decide
(with the help of those lists) which hypotheses to remove from the context.  In
auto mode, the system will perform an automatic retraction if the culprit list
contains a single hypothesis.  Otherwise it reverts to manual.  In either case,
any unresolved inconsistencies are dealt with manually. 

\subsection{Culprit List Algorithm}
Processing the three lists above to create a culprit list ($CL$) results in the smallest, non-empty intersection:
$CL = Min-not-\emptyset \linebreak[0] ((LB \cap MC \cap FS),  \linebreak[0] (LB \cap MC),  \linebreak[0] (LB \cap FS),  \linebreak[0] (MC \cap FS),  \linebreak[0] LB,  \linebreak[0] MC, FS)$, 
where $Min-not-\emptyset$ is a function that chooses the smallest non-empty set from a
list of sets in decreasing order of importance (for tie-breaking purpose---e.g. choose $LB$ over $MC$).

Other factors being equal, removing a belief with low credibility (constraint 4) is currently
preferred over one whose removal  does the least damage (constraint 3), because
credibility orders are not weakened by the absence of deductive closure.  Regarding the information
used to support constraint 3, the system prefers to remove a more common
belief  over one with the fewest supported
nodes, because that will protect beliefs that have been derived in more
ways.  For example, given 
\verb|A, B, D, A->C, B->C, and D->~C,| and the derivations of \verb|C| (both
ways) and \verb|~C|,
the minimally inconsistent sets would be 
\verb|{A, A->C, D, D->~C}| and \verb|{B, B->C, D, D->~C}| 
and both \verb|D| and \verb|D->~C| would be the most common hypotheses. Removal
of either results in the loss of \verb|~C| and the retention of \verb|C|, which
was derived two ways.

It should be emphasized, however, that the final determination of CL is the
smallest, non-empty set found.  In this sense, condition 3 regains some of its
lost status. For example: if $(LB \cap FS)$ contained three hypotheses and $(MC \cap FS)$
contained two, the latter would be chosen over the former, even though only the
former contained credibility information.

An edited sample of the autoBR output for a belief revision exercise is
available in Appendix 1.

\section{Conclusion and Future Work}
By considering the work of coherence and theoretical researchers, we were able
to improve our existing belief revision system by adding information essential
to improving culprit selection during revision. The revised system
combines the constraints of minimal information loss and maximal credibility in
its development of a culprit list during belief revision to return consistency
to a knowledge base.  It considers (a) the number of hypotheses and derived
propositions that will be affected by the revision as well as (b) the relative
credibility orderings of the hypotheses under consideration.  These orderings
are determined by both source credibility orderings and credibility orders
directly assigned between hypotheses.  All source and credibility information
can be retracted, altered, added to, and reasoned about while the system is
running.  

Automatic belief revision is possible when the culprit list contains a single
hypothesis.  In this case, the system appears to the user to be non-monotonic:
i.e. the user can add a proposition to an existing context, but end up with a
belief space that is not a superset of the original belief space.

The issue of incorporating source information and credibility ordering into a
knowledge base is key to maintaining the credibility of an information fusion
system. Selection of a well-believed hypothesis as the culprit (due to other
considerations like minimizing damage to the knowledge base) might also
indicate a need to re-evaluate the reliability of its source.  This might lead
to development of a system that dynamically adjusts source and propositional
credibility orders based on past performance.

Work for the immediate future will include dealing with (1) a proposition
having multiple sources and (2) revising the inconsistent sets as a group.  For
the latter, we might first partition the inconsistent sets based on which of
the contradictory nodes each hypothesis supports (or if it supports both), then
analyzing the groups by their inconsistent set as well as by their partition. Then the system
could better determine which of the contradictory nodes should be contracted
and remove its supports efficiently.  We hope that one result will be an
improvement on safe contraction \cite{AlcMak85}. For example:   If retracting a
proposition \verb|P| with the two origin sets of \verb|{A,B,C}| and \verb|{B,D}| ordered in
increasing credibility, then retracting the least-believed in each set (as per safe-contraction) would
remove both \verb|A| and \verb|B|, when removal of \verb|B| would be
sufficient. This improvement might show up as the union of the least believed
and the most common sets. 

\section{Acknowledgements}
The authors appreciate the insights and feedback of Bill Rapaport, Haythem
Ismail, and the SNePS Research Group.  

\section{Appendix 1}
Below is a description of the information given to the knowledge base.  \texttt{WFF22} is added last with forward inferencing, which produces the contradictory propositions that trigger SNeBR. 
The five sources with their credibility orderings are:  
\begin{tabbing}
\(Holybook>Prof\) \= \verb|WFF1:  GREATER(HOLYBOOK,PROF)|\\
\(Prof>Nerd\)  	\> \verb|WFF2:  GREATER(PROF,NERD)|\\
\(Nerd>Sexist\) \> \verb|WFF3:  GREATER(NERD,SEXIST)|\\   
\(Fran>Nerd\) \> \verb|WFF4:  GREATER(FRAN,NERD)|	
\end{tabbing}

\begin{description}
\item[Source:	statement]
\item[Nerd:  	Jocks aren't smart.]\mbox{} \linebreak[3]\verb|WFF10:  all(X)(JOCK(X) => (~SMART(X)))|
		\linebreak[3]\verb|WFF11:  SOURCE(NERD,WFF10)|
\item[Sexist: 		Females aren't smart.]\mbox{}	\linebreak[3]\verb|WFF12:  all(X)(FEMALE(X) => (~SMART(X)))|
		\linebreak[3]\verb|WFF13:  SOURCE(SEXIST,WFF12)|
\item[Prof: 		Grads are smart.]\mbox{}
\linebreak[3]\verb|WFF14:  all(X)(GRAD(X) => SMART(X))|
			\linebreak[3]\verb|WFF15:  SOURCE(PROF,WFF14)|
\item[HolyBook: 	Old people are smart.]\mbox{} 	\linebreak[3]\verb|WFF16:  all(X)(OLD(X) => SMART(X))|
			\linebreak[3]\verb|WFF17:  SOURCE(HOLYBOOK,WFF16)|
\item[Fran: 	I'm an old, female, jock who's a grad.] \mbox{}
		\linebreak[3]\verb|WFF22:  FEMALE(FRAN) and OLD(FRAN) and 
GRAD(FRAN) and JOCK(FRAN)| 
\linebreak[3]\verb|WFF23:  SOURCE(FRAN,WFF22)|
\end{description}

The following code is an edited version of the system output showing the inconsistencies found and the hypotheses removed through autoBR. Author's comments are in \textit{italics}.
\begin{tabbing}
\end{tabbing}

{\itshape After all the information is in, one  contradiction is detected:}

{\ttfamily The contradiction involves the newly derived proposition:}
\begin{tabbing}
in \= mo \= lots more \= \kill
\> \>    \verb|WFF24:  SMART(FRAN)|
\end{tabbing}
\texttt{and the previously existing proposition:}
\begin{tabbing}
in \= mo \= lots more \= \kill
\> \>    \verb|WFF25:  ~SMART(FRAN)| \\ 
\end{tabbing}

{\itshape To resolve the contradiction, SNePS Belief Revision (SNeBR) analyzes
the inconsistent set formed by the union of  the  two  Origin  Sets for the
contradictory propositions:} 
\begin{tabbing}
in \= mo \= a bunch more space \= \kill  
\>   \textit{(WFF16,WFF22) $\cup$ (WFF12,WFF22) $=$} \\
\> \> \> \textit{(WFF22,WFF16,WFF12)} \\
\end{tabbing}

\textit{The three sets aiding culprit selection:}
\begin{tabbing}
in \= mo \= lots more \= \kill  
\> \texttt{The least believed hypothesis:} \\	   
\> \>   \texttt{(WFF12)}\\
\>  \texttt{The most common hypotheses:}  \\  
\> \>   \texttt{(WFF22 WFF16 WFF12) }\\
\>  \texttt{The hypotheses supporting the fewest nodes:} \\   
\> \>  \texttt{(WFF12 WFF16) }\\
\end{tabbing}

\begin{tabbing}
in \= mo \= lots more \= \kill  
\textit{The system informs the user of its decision:}\\
\> \verb|I will remove the following node:|  \\
\>\> \verb|WFF12:  all(X)(FEMALE(X) => (~SMART(X)))|  \\
\end{tabbing}

{\itshape The system continues reasoning and discovers the same contradiction (derived in new ways):}

{\ttfamily The contradiction involves the newly derived proposition:}
\begin{tabbing}
in \= mo \= a bunch more space \= \kill  
\> \>     \verb|WFF25:  ~SMART(FRAN)|
\end{tabbing}

\texttt{and the previously existing proposition:}
\begin{tabbing}
in \= mo \= a bunch more space \= \kill  
\> \>    \verb|WFF24:  SMART(FRAN) |\\
\end{tabbing}

{\itshape There are two known-to-be-inconsistent sets in the context, now:}

{\ttfamily The following sets are known to be inconsistent. To make the
context consistent, remove at least one hypothesis from each of the sets:}
\begin{tabbing}
in \= mo \= a bunch more space \= \kill  
\>    \>   \texttt{(WFF22 WFF16 WFF10)	(WFF22 WFF14 WFF10)} \\
\end{tabbing}

{\itshape The three sets aiding culprit selection:}
\begin{tabbing}
in \= mo \= a bunch more space \= \kill  
\>   \texttt{The least believed hypothesis:  } \\    
\> \> \texttt{(WFF10) }\\
\>   \texttt{The most common hypotheses:      }\\
\> \> \texttt{(WFF22 WFF10) 		}\\
\>   \texttt{The hypotheses supporting the fewest nodes:    }\\
\> \> \texttt{(WFF14 WFF10) }\\
\end{tabbing}

{\itshape The system's decision in this case: }
\begin{tabbing}
in \= mo \= a bunch more space \= \kill  
\>   \texttt{I will remove the following node:}\\
  \> \>    \verb|WFF10:  all(X)(JOCK(X) => (~SMART(X)))| \\
\end{tabbing}

\bibliography{bibliography2} 
\bibliographystyle{aaai}
\end{document}